%% file: acl_latex.tex
\pdfoutput=1

\documentclass[11pt]{article}

\usepackage[final]{acl}

\usepackage{times}
\usepackage{latexsym}

\usepackage[T1]{fontenc}

\usepackage[utf8]{inputenc}

\usepackage{microtype}

\usepackage{inconsolata}

\usepackage{graphicx}
\usepackage{tabularx}
\usepackage{booktabs} 
\usepackage{multirow} 
\usepackage{array}    
\usepackage{makecell}
\usepackage{bbding}
\usepackage{multirow}
\usepackage{pgfplots}
\usepackage{svg}
\usepackage{amsmath} 
\usepackage{amssymb}
\usepackage{url}
\usepackage{tablefootnote}
\usepackage{graphicx}
\usepackage{cuted}
\usepackage{tabularx}
\usepackage{xcolor}
\usepackage{pifont}
\usepackage{tcolorbox}
\usepackage{pgfplots}
\pgfplotsset{compat=1.18}
\usepackage{tikz}
\usepackage[table,xcdraw]{xcolor}
\tcbuselibrary{breakable, skins}

\newcommand{\datasetname}{PEC-Home}

\definecolor{deepgreen}{RGB}{6,153,6}  
\definecolor{deepred}{RGB}{254,34,35}  
\definecolor{deeporange}{RGB}{255,140,0}
\definecolor{darkblue}{RGB}{0,0,139}
\definecolor{roomcolor}{RGB}{80,114,196}
\definecolor{devicecolor}{RGB}{237,125,49}
\definecolor{operationcolor}{RGB}{112,173,100}
\definecolor{paramcolor}{RGB}{234,107,102}
\definecolor{exampleorange}{RGB}{255,242,204}
\definecolor{examplegreen}{RGB}{34,139,34}
\definecolor{myorange}{HTML}{f4bebe}
\definecolor{mypurple}{HTML}{bebef4}

\title{\datasetname: Interpretation of Progressively Elliptical Commands in\\ Smart Homes}

\author{%
Yingyu Shan\textsuperscript{1},
Zeming Liu\textsuperscript{2},
Silin Li\textsuperscript{1},
Boao Qian\textsuperscript{1},\\
\textbf{Jiashu Yao}\textsuperscript{1},
\textbf{Yuhang Guo\textsuperscript{1\dag},} 
\textbf{Haifeng Wang\textsuperscript{3}} \\
\textsuperscript{1}Beijing Institute of Technology \quad
\textsuperscript{2}Beihang University \quad
\textsuperscript{3}Baidu Inc. \\
\textsuperscript{\dag}Corresponding author \quad
Email: \texttt{shanyingyu@bit.edu.cn, guoyuhang@bit.edu.cn}
}

\begin{document}
\maketitle
\begin{abstract}

Recent advancements in Large Language Models (LLMs) have empowered home assistants with natural language interaction capabilities. However, current assistants overlook the progressive omission that occurs in human dialogue as shared context accumulates, leading to more elliptical expressions for efficient communication. Thus, current assistants still struggle to interpret such elliptical expressions accurately, which limits their effectiveness in real-world applications. In practical smart home scenarios, assistants face two major challenges caused by elliptical commands: (1) referential ambiguity caused by different environmental expectations among multiple users; and (2) intention ambiguity resulting from user preferences that evolve over time or change with the environment. To address these challenges, we introduce \datasetname, the first simulated home dataset specifically designed for interpreting progressively elliptical commands in smart homes. Extensive experiments on various LLMs, including GPT-4o, show that existing home assistants struggle to execute user-intended operations based solely on elliptical commands. Even when equipped with tools for storing and retrieving user dialogue history, execution accuracy remains below that achieved with complete commands. Our code and dataset are available at \url{https://github.com/BITHLP/PEC-Home}.
\end{abstract}

\begin{figure}[t]
    \centering
    \includegraphics[width=\linewidth]{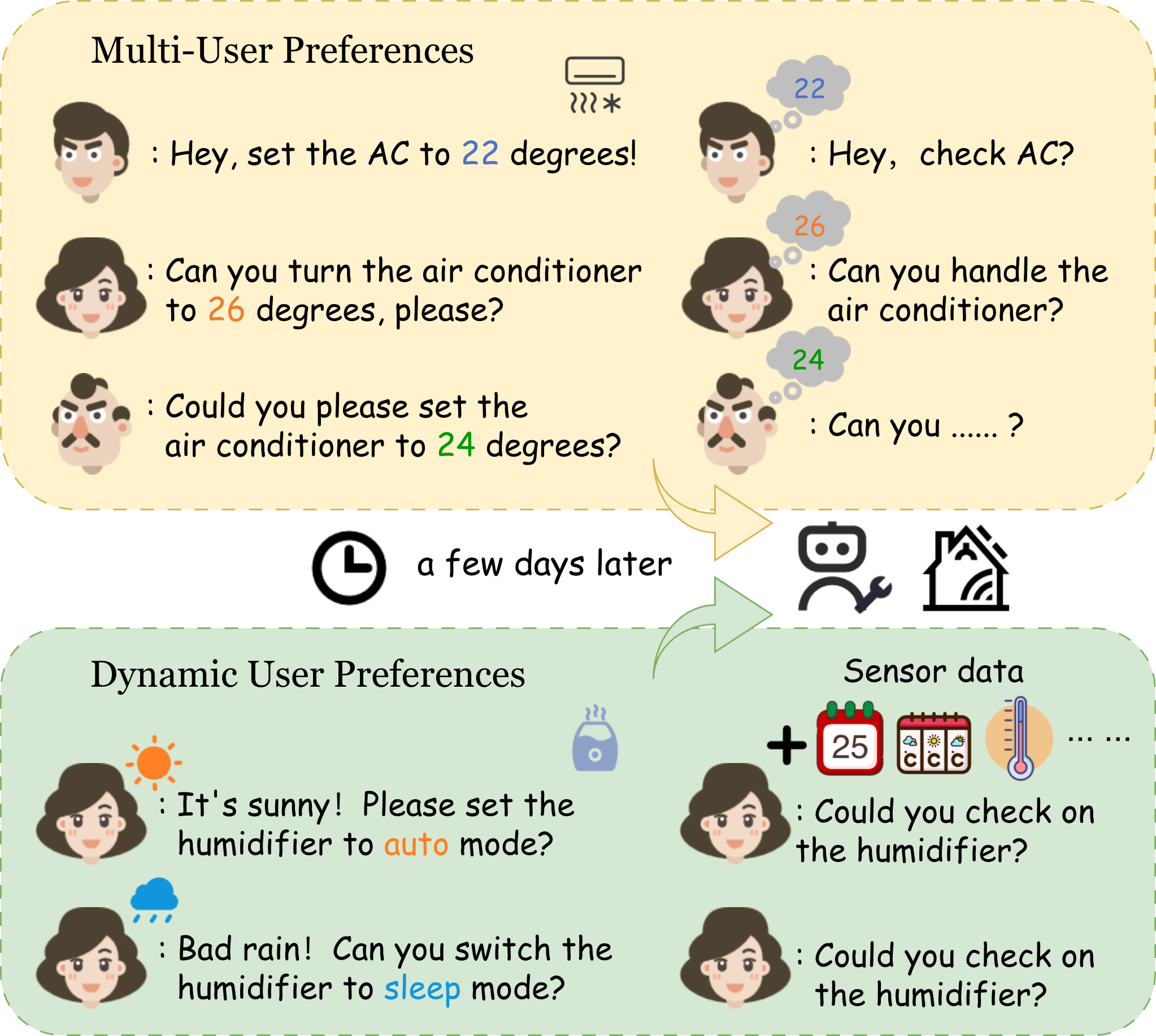}
    \caption{\label{figure:example}An example of \datasetname. \textcolor{deeporange}{Multi-User Preferences} presents the referential ambiguity from conflicting "comfortable temperature" definitions between family members and \textcolor{examplegreen}{Dynamic User Preferences} indicates the intention ambiguity caused by environment changes.}
\end{figure} 
\section{Introduction}
Home assistants automate routine household tasks and enhance user interaction through intuitive, context-aware support, seamlessly integrating into daily life \citep{10.1145/329124.329126}. Specifically, their automation leads to increased convenience \cite{ur2014practical}, optimized energy usage \citep{sepasgozar2020systematic, 9288505} and so on. 

In the pre-LLM era, home assistants primarily focused on executing predefined rules and pattern recognition tasks like household routine automation \citep{dey2006icap, ur2014practical} and activity prediction \citep{tax2018human, kim2017activity, khraief2019convolutional}. While these assistants achieved home automation, they fundamentally lacked the capability to interact with users through natural language.

The advent of LLMs \citep{vaswani2017attention} revolutionized home assistants by enabling natural language understanding across domains \citep{OpenaiGPT4, touvron2023Llama2, deepseekai2024deepseekv3technicalreport}. Recent LLM-based home assistants like Sasha \cite{10.1145/3643505} and SAGE \cite{rivkin2024aiot} demonstrate improved device control capability facing commands like "Make it less chilly". However, these LLM-based assistants operate under static, single-user assumptions or oversimplify dynamic user command shifts as merely explicit or ambiguous states, disregarding the progressive shift from explicit to elliptical through long-term interactions. Cognitive science researches have shown that humans naturally develop idiosyncratic and increasingly elliptical conventions with familiar partners over repeated interactions \citep{krauss1964changes, zwaan1998situation, hawkins2020characterizing}. By neglecting this phenomenon, current home assistants fall short in handling naturally elliptical commands, preventing users from interacting with them as efficiently and intuitively as they would with human partners. This limitation fundamentally undermines home assistants' ultimate goal of providing intuitive, context-aware support.

To address this limitation, we introduce \datasetname, the first simulated home dataset that is specifically designed for simulating progressively elliptical commands in smart homes. 
The interpretation of progressively elliptical commands encounters two core challenges in practical scenarios (as shown in Figure \ref{figure:example}): referential ambiguity arising from conflicts in multi-user preferences (e.g., conflicting "comfortable temperature" definitions between family members), and intention ambiguity in dynamic user preferences, where previously explicit commands (e.g., "set the humidifier to auto mode") gradually lose specificity as the environment changes.
\datasetname~comprises 1,780 dialogues from 1,424 personas \citep{zhang-etal-2018-personalizing}, providing a novel and more practical perspective for enhancing and evaluating the performance of LLM-based home assistants.

To evaluate the effectiveness current LLM-based home assistants in interpreting progressively elliptical commands, we conduct extensive experiments on \datasetname. We investigate a range of LLMs using both zero-shot prompting and in-context learning to assess their ability to execute user-intended operations based solely on elliptical commands. The results show that none of the evaluated LLMs are able to reliably interpret these elliptical instructions. Furthermore, we enhance LLMs with external tools, retrieval-augmented generation (RAG) for accessing dialogue history, or fine-tuning LLMs aimed at improving command interpretation capability. However, our experiments demonstrate that even state-of-the-art models such as GPT-4o fail to maintain their performance achieved on complete commands when handling elliptical ones, highlighting the limitations of existing methods.

\begin{itemize}
    \item To the best of our knowledge, we are the first to identify the task of progressively elliptical commands interpretation in human–home assistant interactions.
    \item To facilitate the study of this task, we introduce \datasetname, the first simulated home dataset modeling the progressive shift from explicit to elliptical commands.
    \item Our experimental results on 10 distinct LLMs demonstrate that all models experience substantial performance drops when interpreting progressively elliptical commands. Even with enhancements such as tool integration, RAG, and fine-tuning, these models including GPT-4o still fail to achieve reliable execution accuracy on highly elliptical commands.
\end{itemize}

\section{Related Work}
\subsection{Pre-LLM Home Automation Systems}
Home assistants in the pre-LLM era were essentially home automation systems that relied on rule-based or machine learning algorithms. Among the earliest approaches, \citet{dey2006icap} proposed iCAP, a rule-based system enabling users to create rules to automate home devices. Similarly, \citet{ur2014practical} explored trigger-action programming, enabling users to create custom automation rules (e.g., "If it is 6 p.m., then turn the lights on").

The advancement of machine learning has enabled home automation systems to integrate diverse algorithms, including SVM for emotion-based automation \cite{jaihar2020smart}, CNN for elderly fall detection \cite{khraief2019convolutional}, and LSTM for next-activity prediction in multi-user environments \cite{kim2017activity}.
\citet{9612040} investigated how RL-based smart homes can influence human behavior. Similarly, \citet{9288505} proposed a multi-objective RL framework aiming to optimize power consumption in smart homes. 

\datasetname~fundamentally differs from pre-LLM automation systems by shifting from rule-based automation and single-task pattern recognition to natural language driven progressive elliptical commands resolution.

\begin{figure*}[!ht]
    \centering
    \includegraphics[width=\linewidth]{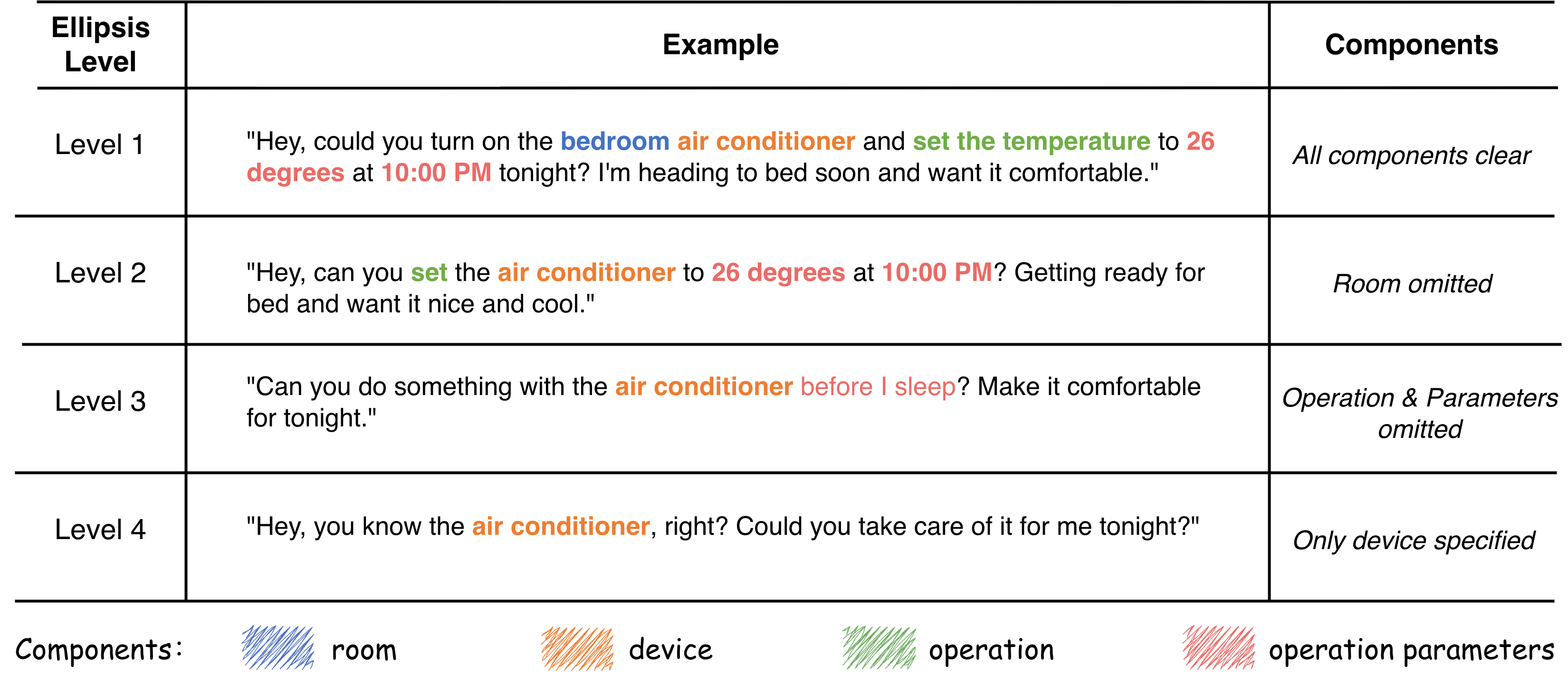}
    \caption{\label{figure:cases}
    Examples of progressively elliptical user commands across four levels (Lv1–Lv4) illustrate the defined standards and the systematic omission of four core components (room, device, operation, and operation parameters). We color \textcolor{roomcolor}{room}, \textcolor{devicecolor}{device}, \textcolor{operationcolor}{operation}, and \textcolor{paramcolor}{operation parameters}. We emphasize that the examples in the figure are simplified for better understanding. 
    }
\end{figure*} 

\subsection{LLM-Based Home Assistants}
Humans often bridge abstract concepts (e.g., `comfort') and device-specific actions (e.g., `turning on the air conditioner') through intuitive semantic associations \cite{10.1145/3643505}. LLMs demonstrate the ability to understand the underlying concrete actions intended by humans \cite{he2024can}. This capability has inspired recent works to integrate LLMs in resolving ambiguous commands.
 
Sasha \cite{10.1145/3643505}, an LLM-based smart home assistant that generates action plans for under-specified user commands through iterative reasoning. \citet{rivkin2024aiot} proposes an LLM-based agent SAGE that dynamically generates prompt trees and toolchains to flexibly handle user requests. \citet{yin2024harmonyhomeagentresponsive} introduces Harmony, a framework leveraging the locally deployable LLM to address user needs. AwareAuto \cite{shi2024bridginggapnaturaluser}, the first end-user programming system leveraging LLMs to bridge user expressions with smart home automation. 

In contrast to prior research focus on ambiguous commands, our work identifies the task of progressively elliptical commands interpretation, which is intuitively emerging through long-term interactions.

\section{Problem Definition}
The ultimate task of home assistants is to interpret a user's command $u_t$ at turn $t$ within a smart home environment $H$. Each command, which can be explicit or elliptical, must be accurately mapped to a specific, executable assistant response $a_t$. This response typically takes the form $r_i.d_j.m_k(\theta)$, indicating a target room $r_i$, a device-method $d_j.m_k$, and parameters $\theta$.

While baseline approaches (e.g., prompt-based methods) perform this mapping using only the current command and home state, $f_{\text{baseline}}: (u_t, H) \rightarrow a_t$, the challenge of interpreting elliptical commands necessitates leveraging contextual information. Primarily, this involves the dialogue history, $\mathcal{C} = \{(u_1, a_1), \dots, (u_{t-1}, a_{t-1})\}$, leading to advanced interpretation approaches such as $f_{\text{advanced}}: (u_t, \mathcal{C}, H) \rightarrow a_t$. 

In practical smart home scenarios, assistants face two major challenges caused by elliptical commands: (1) Multi-User Preferences, which primarily involves mapping $f_{\text{multi}}(u_t, \mathcal{C}_l, H) \rightarrow a_t$ using selected user-specific dialogue histories $\mathcal{C}_l$, and (2) Dynamic User Preferences, which adapts to environmental states $\mathcal{C}_l^t$ and sensor data $S_t$ via the function $f_{\text{dyn}}(u_t, \mathcal{C}_l^t, H, S_t) \rightarrow a_t$. We provide more detailed problem definition in the Appendix \ref{baselines details}.

\section{\datasetname}
\datasetname~(\textbf{P}rogressively \textbf{E}lliptical \textbf{C}ommands Dataset for Smart Homes) comprises 1,780 dialogues involving 1,424 unique personas from PersonaChat \cite{zhang-etal-2018-personalizing}. Each dialogue includes a sequence of 4 chats that become progressively more elliptical. \datasetname~is designed to capture and simulate the progressive ellipsis of user commands over collaborative communication, which is a well-studied phenomenon in cognitive science \citep{krauss1964changes, clark1986referring, hawkins2020characterizing}.
To capture different aspects of this progressive shift from explicit to elliptical, \datasetname~is divided into two parts: 1) \textbf{Multi-User Preferences} and 2) \textbf{Dynamic User Preferences}. In this section, we provide a comprehensive description of the collection process of \datasetname, a detailed comparison with existing datasets, and a thorough presentation of its statistics.

\subsection{Dataset Collection}

\textbf{Virtual Environment Construction} 
We constructed a virtual home environment that includes 12 types of devices, covering common device categories such as lighting, humidifiers, etc. Each device is equipped with multiple personalized executable methods, totaling over 50 distinct methods. A comprehensive list of all devices and their executable methods is provided in the Appendix \ref{sec:appendix_env}. To simulate real-world scenarios, these devices are distributed across multiple rooms in our virtual environment. A detailed list of devices in each room is available in the Appendix \ref{sec:appendix_env}. In total, more than 350 personalized methods are allocated to different rooms, ensuring diversity and comprehensive scenario coverage in our dataset.

\input{latex/tables/comparsion}
\input{latex/tables/statistics}

\textbf{User Commands Generation} 
Based on this virtual environment, we first generated function-call style device operations.
Based on above, we utilized unique personas for each user and generated personalized parameters aligned with their preferences. For multi-user preferences, we randomly selected three distinct users to form a household. For dynamic user preferences, we generated two preference parameters for each user within an device method, each corresponding to different environmental states. We ensured that all operations included memorable personalized parameters.

The increasing ellipsis in PEC-Home's user commands is designed to reflect patterns observed in real human communication. Prior study \citep{krauss1964changes} have shown that when interlocutors repeatedly refer to the same object, their references becomes more elliptical in collaborative process. Initial detailed descriptions are gradually shortened by omitting elements which mutually established within the shared context, converging on brief labels. To capture this progressive shift from explicit to elliptical commands in user-assistant interactions, PEC-Home's ellipsis levels are constructed by gradually reducing the amount of explicit information in commands. Inspired by \citet{10.1145/3643505}, which identifies four core components of smart home commands:\textbf{room}, \textbf{device}, \textbf{operation}, and \textbf{operation parameters} (e.g., `brightness set to 3'), we define four ellipsis levels by omitting these components. To ensure data quality and minimize manual effort, we utilize GPT-4o to generate natural language commands based on function-call style operations and user personas, ensuring clarity, grammatical accuracy, and alignment with casual conversation style. The four defined levels are:
\begin{itemize}
    \setlength{\itemsep}{0em}
    \setlength{\parskip}{0pt}
    \setlength{\parsep}{0pt}
    \item \textbf{Level 1:} The user command includes all four core components.
    \item \textbf{Level 2:} 
    As fundamental contextual elements such as time and space are usually established before introducing events in human communication \citep{zwaan1998situation}, we assume users first establish spatial location during interactions. Following prior work on conceptual pacts and common ground \citep{brennan1996conceptual, clark1986referring}, once this spatial reference is mutually established through previous commands, repeating it for subsequent commands within the same space becomes redundant. Thus, we omit the \textbf{room} information at this level.
    \item \textbf{Level 3:} Previous research \citep{carroll1980naming} has found that descriptive modifiers in user language are gradually omitted over time. After removing location information, the remaining modifiers mainly include explicit \textbf{operation} and \textbf{operation parameters}. Therefore, we omit these descriptive components at this level.
    \item \textbf{Level 4:} Following the findings that interactions often converge towards idiosyncratic, shared references \citep{krauss1964changes}, commands at this level only explicitly specify the \texttt{device}, simulating the ultimate reference collaborators have achieved.
\end{itemize}
The prompt used to guide LLMs in generating commands at each level was manually reviewed and adjusted to ensure generated commands met the defined standards. Detailed examples of progressively elliptical commands are provided in Figure \ref{figure:cases}, and specific prompts used to guide GPT-4o in generating commands are detailed in Appendix \ref{sec:appendix_cmd}.

\textbf{Quality Assessment} 
Inspired by \citet{zeng2024fame}, we apply manual sampling and validation procedures to validate the reliability of our operation generation process. Three graduate researchers conducted a manual evaluation of 500 randomly selected function-call style device operations. The results demonstrated that 100\% of the operations were correct, and 96\% were properly aligned with environmental states, highlighting the high accuracy and reliability of our operation generation approach. To confirm the credibility of our user commands generation framework, the same three researchers manually assessed 200 dialogues which involving 
multiple progressively elliptical commands. The results revealed that 94.5\% of these commands met the defined standards, demonstrating the reliability of the entire process.

\subsection{Comparison}

Table \ref{table:data_comparison} compares \datasetname~with existing simulated home datasets, highlighting its unique contributions. Unlike existing datasets, \datasetname~is specifically designed to  model the shift from explicit to elliptical that naturally arises from long-term human-assistant interactions. In practical smart home scenarios, interpreting such progressively elliptical commands presents two core challenges (as illustrated in Figure \ref{figure:example}). To address these, \datasetname~not only captures these ambiguities but also provides personalized long-term interaction data to facilitate the study of progressively elliptical commands interpretation.

\subsection{Statistics}
Table \ref{table:Statistics} summarizes the statistics for the \datasetname~dataset across two tasks: Multi-User Preferences and Dynamic User Preferences. Both tasks show a clear trend where the average token count decreases from Lv1 to Lv4, indicating that the user commands become progressively shorter and elliptical. Additionally, the percentages of Room, Device, and Operation components also decrease as the commands become more elliptical, further supporting that the commands are becoming more elliptical and less specific.

The lack of statistics for \textbf{Operation parameters} component is due to the presence of highly common parameters, such as ``up'', ``sleep'', etc. These words frequently appear in user commands making them difficult to track accurately. Additionally, time-related parameters often appear in the 12-hour format (e.g., "11:30 PM") rather than the 24-hour format, further complicating the precise identification of time-related parameters. As a result, we didn't provide statistics for the operation parameters component.

\input{latex/tables/main}

\section{Experiment}
\subsection{Setup}

\textbf{Models}
\label{sec:Models}
We select several open-source and closed-source LLMs for a comprehensive evaluation \citep{zhuang2023toolqa, wang-etal-2024-appbench}. Specifically, the open-source models including: Llama-3.1-8B-Instruct \cite{llama3modelcard}, Mistral-7B-Instruct-v0.3 \cite{jiang2023mistral7b}, Google's Gemma series (gemma-2-9b/27b-it) \cite{gemmateam2024gemma2improvingopen}, the Qwen2.5 Series (Qwen2.5-7B/14B/32B/72B-Instruct) \cite{Qwen2.5}, and the DeepSeek-V3 \cite{deepseekai2024deepseekv3technicalreport}. For closed-source models, we choose GPT-4o \cite{OpenaiGPT4}.  We exclude reasoning models as LLM-based assistants require rapid responses, and such models typically introduce delays by first generating reasoning trajectory.

\textbf{Methods}
We experiment with Zero-Shot Prompting, In-Context Learning, and Retrieval-Augmented Generation on LLMs mentioned above. For specific implementation details, please refer to the Appendix \ref{appen_exp}.

\subsection{Metrics}
Following previous LLM-based home assistants \citep{shi2024bridginggapnaturaluser, rivkin2024aiot} and code generation tasks \citep{chen2021evaluating, yu2018spider}, we use \textbf{Execution Accuracy (EA)} as the primary metric to evaluate the performance of the home assistant in executing user instructions. Specifically, the correct execution of the home assistant is defined as generating the precise device control operations required to accurately fulfill the user's command within our virtual environment.

To evaluate the operational accuracy of home assistants in executing operations accurately, we utilize the \textbf{F1} score \citep{devlin-etal-2019-bert, wang-etal-2024-order}. The specific formula for calculating F1 is provided in the Appendix \ref{app_metrics}.

\subsection{Results}
Table \ref{table:main_result} shows the results of the experiments on \datasetname. Several conclusions can be drawn from the results.

\textbf{The results indicate that no models can accurately execute user-intended operations based solely on elliptical commands}. Even when augmented with RAG to access dialogue history, the performance of these models on elliptical commands does not match that achieved with complete commands. Specifically for RAG methods, while dialogue history offers some assistance, the performance in correctly executing user operations still declines as the commands become more elliptical.

\textbf{While increased model parameters can enhance a model's ability to leverage dialogue history for command interpretation, this advantage sharply declines when addressing highly elliptical commands.} Experimental results demonstrate that although larger models (e.g., Qwen2.5 and Gemma2 series) show improved performance on less elliptical commands (Lv2), these gains rapidly diminish facing highly elliptical commands. Even SOTA model like GPT-4o, under In-Context Learning, underperforming some smaller RAG-equipped models on Lv4 commands. This emphasizes that \textbf{merely increasing parameter size is not a sufficient strategy for accurately executing user-intend operations facing elliptical commands.}

\textbf{EA and F1 reveal divergent failure modes across ellipsis levels.} On less elliptical commands, EA is often higher than F1. This discrepancy is primarily attributed to two factors: firstly, as detailed in Appendix \ref{app_metrics}, EA does not require output format to be strictly correct, unlike F1, which highlights the current limitations of LLMs in adhering to precise required formats. Secondly, models may exhibit "over-execution", where they correctly complete the core instruction but also generate extra operations, thereby lowering the F1 score but not affecting the EA. Conversely, on highly elliptical commands, EA tends to be lower than F1. This phenomenon often occurs because models take "shortcuts". For instance, in response to a command like "turn on the light and set brightness to 4," a model might only generate the simpler "turn\_on" operation. Such partially matched simple operations can boost F1 scores, but the EA score remains low, leading to the observed divergence between these two metrics.


\section{Analysis}
In this section, we extend our analysis by conducting three additional research questions (RQs) to further investigate elliptical command interpretation. These RQs are designed to systematically explore the key factors that influence the assistants' ability to resolve elliptical command, focusing on \textbf{memory management (RQ1)}, \textbf{model optimization (RQ2)}, and \textbf{external tools integration (RQ3)}. For experiment details, please refer to Appendix \ref{appen_exp} and a comprehensive error analysis is provided in Appendix \ref{Error Analysis}.

\begin{figure}[!ht]
    \centering
    \includegraphics[width=\linewidth]{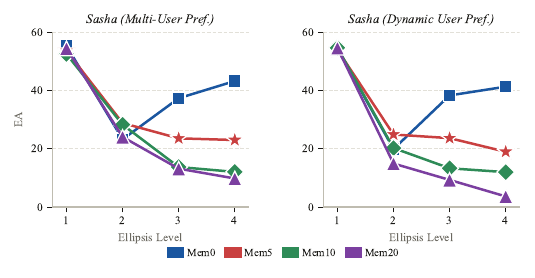}
    \caption{\label{sasha_mem} Execution Accuracy of Sasha on Qwen2.5-7B-Instruct across varying amounts of preloaded memory in multi-user preferences and dynamic user preferences scenarios. `Mem number' indicates the amount of preloaded irrelevant memory.}
\end{figure} 

\subsection{RQ1: How Does Preloaded Irrelevant Memory Affect Interpretation Accuracy?}
In practical applications, home assistants often need to handle a large number of commands, and external memory tools are considered to enhance model performance \citep{NEURIPS2020_6b493230, gui-etal-2022-kat}. However, the effectiveness of utilizing memory is influenced by the amount of preloaded irrelevant memory. Figure \ref{sasha_mem} shows the EA of Sasha and Figure \ref{SAGE_mem}, \ref{RAG_mem}, and \ref{RAG_mem_gemma} in Appendix show the EA of SAGE, RAG (Qwen2.5-7B) and RAG (Gemma2-9B) across varying amounts of preloaded memory.

This section investigates how the amount of preloaded irrelevant memory impacts interpretation performance. For Sasha when no irrelevant memory is present in its database, Sasha can effectively leverage dialogue history to enhance its performance on highly elliptical commands. However, when irrelevant information is preloaded into the database, a general trend is observed across all evaluated methods that their EA declines as the commands become more elliptical. This phenomenon underscores the challenges current methods face in filtering and utilizing relevant history dialogue from a noisy memory environment when interpreting increasingly elliptical commands.

\begin{figure}[!ht]
    \centering
    \includegraphics[width=\linewidth]{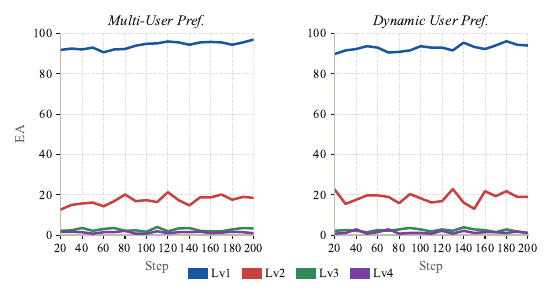}
    \caption{Execution Accuracy of Qwen2.5-7B-Instruct model in
    test dataset under different training steps.}
    \label{ft}
\end{figure} 

\subsection{RQ2: Does Fine-tuning Enhance Elliptical Commands Resolution?}
Fine-tuning is a common approach to adapting LLMs to specific domains. We split the dataset into training, validation, and test sets in a 5:1:4 ratio, and then used the training dataset to fine-tune the Qwen2.5-7B and Gemma2-9B models. These models are selected as our main experiments involve evaluations of both the Qwen and Gemma model series. Figure \ref{ft} shows the results on Qwen2.5-7B and figure \ref{ft_gemma} in Appendix shows the results on Gemma2-9B.

We use the same prompt as the one used for the ICL method to guide fine-tuned model inference. The results show that while fine-tuning significantly boost performance on low-ellipsis, its performance collapse on highly elliptical commands. This performance on highly elliptical tasks mirrored that of ICL methods in Table \ref{table:main_result}, demonstrating that fine-tuning, despite aiding in interpreting complete commands, fails to resolve the core challenge of interpreting progressively elliptical commands.

\begin{table}[!ht]
    \centering
    \begin{tabular}{ccccc}
      \hline
        \multirow{2}*{\textbf{Method}} & \multicolumn{4}{c}{\textbf{Multi-User Preference}} \\
    \cline{2-5}
        ~ & Lv1 & Lv2 & Lv3 & Lv4 \\
      \hline
         SAGE   & 51.40 & 30.25 & 15.73 & 4.49 \\
         Sasha  & 55.34 & 23.25 & 37.36 & 43.26 \\
      \hline
        \multirow{2}*{\textbf{Method}} & \multicolumn{4}{c}{\textbf{Dynamic User Preference}} \\
    \cline{2-5}
        ~ & Lv1 & Lv2 & Lv3 & Lv4 \\
      \hline
         SAGE   & 52.00 & 34.55 & 18.67 & 5.33 \\
         Sasha  & 54.33 & 19.93 & 38.33 & 41.33 \\
      \hline
    \end{tabular}
    \caption{Execution Accuracy of SAGE and Sasha on Qwen2.5-7B-Instruct across varying levels of ellipsis in multi-user preferences and dynamic user preferences scenarios.}
    \label{tab:tools}
\end{table}

\subsection{RQ3: How Do External Tools Improve Ellipsis Command Handling?}
The integration of external tools has become mainstream in the current home assistants and autonomous agents \citep{qiao-etal-2024-autoact, chen-etal-2024-eval,zhang-etal-2024-codeagent}. 
For instance, \textbf{Sasha} and \textbf{SAGE} represent two mainstream approaches among current LLM-based assistants. For specific implementation details, please refer to the Appendix \ref{baselines details}.
Thus, this section explores the performance of current tool-based home assistants in \datasetname. Table~\ref{tab:tools} shows the results. 

Both Sasha and Sage fall short compared to ICL approaches facing complete commands. This suggests that tool-based methods struggle with basic user commands. Among Lv2 commands, where instructions are moderately elliptical, Sasha fails to decide whether invoke external memory, leading to lower EA scores compared to more elliptical commands. In contrast, SAGE exhibits consistently poor performance across all command levels.
These results show that while tool-based methods can handle some elliptical commands, but they fail to consistently resolve elliptical commands. 


\section{Conclusion}
We identify the task of interpreting progressively elliptical commands in smart homes, which is naturally arises from human communication and represents a well-studied phenomenon in cognitive science. To facilitate the study of this crucial task, we introduced \datasetname, the first simulated home dataset specifically designed for interpreting such progressively elliptical commands. Our experimental results demonstrate a significant challenge for current LLM-based assistants. All evaluated models struggle to accurately execute user-intended operations based solely on elliptical commands. Furthermore, even augmented with advanced techniques such as RAG, external tools, and fine-tuning, these LLMs, including state-of-the-art models like GPT-4o, still exhibit substantial performance degradation as commands become more elliptical compared to their performance when facing complete commands.


\section*{Limitations}
Our virtual environment encompasses a diverse range of devices and each device is associated with numerous methods. These methods often come with detailed descriptions. When these descriptions are provided as input to the model, the context becomes very long, leading to higher computational costs. 

Furthermore, due to strict privacy constraints inherent in smart home environments, we are unable to collect large-scale, real-world user interaction data. While real production data is the ideal standard, large-scale access is currently unfeasible. To mitigate this limitation, we utilized LLMs for data generation but took extensive measures to ensure the simulated dialogues faithfully reflect real human communication habits. Prior to dataset construction, we conducted a pilot study on real human-home interactions, observing that users naturally tend to omit room and operation information as shared context grows. These empirical findings, firmly grounded in cognitive science and linguistic theories, directly guided our simulation design. Finally, we implemented a rigorous human-in-the-loop verification process, where experts manually assessed a random sample of 50 dialogues. The results showed that 96.5\% of the generated commands were contextually natural and met human linguistic patterns. Through these efforts, we ensure that PEC-Home serves as a reliable, highly realistic, and privacy-safe proxy for actual human-system interactions.

\section*{Ethical Statement}
During the construction of \datasetname, we adhered to strict ethical standards and ensured that all procedures complied with ethical standards. The virtual environment and the devices within it were manually constructed, carefully reviewed, and meticulously validated to ensure their accuracy and reliability. Following the generation of user commands, we conducted strict quality assessments on the dataset to verify its integrity and ensure it complies with high standards. As a result, we are confident that \datasetname~does not contain any offensive or biased content. Furthermore, all research was performed with a strong commitment to ethical principles, ensuring transparency, and fairness.

\section*{Acknowledgment}
This work is supported by the National Natural Science Foundation of China (Grant No. U21B2009).


\bibliography{custom}

\appendix

\section{Detailed Problem Definition}
\label{Detailed Problem Definition}
Let a virtual smart home environment be represented as:
\begin{equation}
\setlength{\abovedisplayskip}{2pt}
\setlength{\belowdisplayskip}{2pt}
    H = \{r_1, r_2, \dots, r_n\}
\end{equation}
where each room $r_i \in H$ contains operable devices:
\begin{equation}
\setlength{\abovedisplayskip}{2pt}
\setlength{\belowdisplayskip}{2pt}
    D_i = \{d_1, d_2, \dots, d_m\}
\end{equation}

Each device $d_j \in D_i$ is defined by:
\begin{equation}
\setlength{\abovedisplayskip}{2pt}
\setlength{\belowdisplayskip}{2pt}
    d_j = \langle s_j, M_j \rangle
\end{equation}
where $s_j = [s^1_j, \dots, s^p_j]$ is the device state vector and $M_j = \{m^1_j(\theta_1), \dots, m^k_j(\theta_k)\}$ is the executable method set.

Given the current user instruction $u_t$ at turn $t$, the dialogue history is:
\begin{equation}
\setlength{\abovedisplayskip}{2pt}
\setlength{\belowdisplayskip}{2pt}
    \mathcal{C} = \{(u_1, a_1), \dots, (u_{t-1}, a_{t-1})\} \label{eq:dialogue_context_v3}
\end{equation}
where each pair $(u_k, a_k)$ (for $k < t$) represents the user's instruction $u_k$ and the system's corresponding response $a_k$.

Mapping user instruction $u_t$ to system response $a_t$ varies:
\begin{itemize}

    \item Baseline systems (e.g., prompt-based methods) map $u_t$ to $a_t$ using only the home state $H$:
    \begin{equation}
    \setlength{\abovedisplayskip}{2pt}
    \setlength{\belowdisplayskip}{2pt}
        f_{\text{baseline}}: (u_t, H) \rightarrow a_t
    \end{equation}
    \item Advanced systems (e.g., RAG) also use dialogue history $\mathcal{C}$ (Eq. \ref{eq:dialogue_context_v3}) for contextual interpretation of $u_t$:
    \begin{equation}
    \setlength{\abovedisplayskip}{2pt}
    \setlength{\belowdisplayskip}{2pt}
        f_{\text{advanced}}: (u_t, \mathcal{C}, H) \rightarrow a_t
    \end{equation}
\end{itemize}
The model's output $a_t = r_i.d_j.m_k(\theta)$ (target room $r_i$, device-method $d_j.m_k$, parameters $\theta$) is the system's response. The new pair $(u_t, a_t)$ updates the dialogue history.

\subsection{Multi-User Preferences Management}
Let $\mathcal{P} = \{p_1, \dots, p_k\}$ denote the user persona set where each user $p_l$ maintains a distinct dialogue history:
\begin{equation}
\setlength{\abovedisplayskip}{2pt} 
\setlength{\belowdisplayskip}{2pt}
    \mathcal{C}_l = \{(u^l_1, a^l_1), \dots, (u^l_{t-1}, a^l_{t-1})\}
\end{equation}

Advanced systems dynamically select relevant history through:
\begin{equation}
\setlength{\abovedisplayskip}{2pt} 
\setlength{\belowdisplayskip}{2pt}
    \mathcal{C}_l = \mathop{\mathrm{LLM\_Select}}(u_t, \mathcal{P}) \in \{\mathcal{C}_1, \dots, \mathcal{C}_k\}
\end{equation}

The operational mapping then becomes:
\begin{equation}
\setlength{\abovedisplayskip}{2pt} 
\setlength{\belowdisplayskip}{2pt}
    f_{\text{multi}}: (u_t, \mathcal{C}_l, H) \rightarrow r_i.d_j.m_k(\theta)
\end{equation}

\subsection{Dynamic User Preferences Adaptation}
For each user $p_l$, their dialogue history contains two environment-specific preference contexts:
\begin{equation}
\setlength{\abovedisplayskip}{2pt} 
\setlength{\belowdisplayskip}{2pt}
    \mathcal{C}_l = \{\mathcal{C}_l^{e_1}, \mathcal{C}_l^{e_2}\}
\end{equation}
where $e_1$ and $e_2$ represent distinct environmental states. Given real-time sensor data $S_t$, advanced systems dynamically select the relevant context:
\begin{equation}
\setlength{\abovedisplayskip}{2pt} 
\setlength{\belowdisplayskip}{2pt}
    \mathcal{C}_l^t = \mathop{\mathrm{LLM\_Select}}(S_t, \mathcal{C}_l) \in \{\mathcal{C}_l^{e_1}, \mathcal{C}_l^{e_2}\}
\end{equation}

The operational mapping then becomes:
\begin{equation}
\setlength{\abovedisplayskip}{2pt} 
\setlength{\belowdisplayskip}{2pt}
    f_{\text{dyn}}: (u_t, \mathcal{C}_l^t, H, S_t) \rightarrow r_i.d_j.m_k(\theta)
\end{equation}

\begin{table*}[]
\centering
\resizebox{\textwidth}{!}{%
\begin{tabular}{lc}
\hline
\textbf{Room Name} & \textbf{Devices} \\ 
\hline
Master Bedroom & light, air\_conditioner, heating, fan, air\_purifiers, aromatherapy, trash, humidifier, dehumidifiers, tv \\ 
Guest Bedroom & light, air\_conditioner, heating, fan, air\_purifiers, trash, humidifier, dehumidifiers, tv\\ 
Living Room & light, air\_conditioner, heating, fan, air\_purifiers, aromatherapy, trash, humidifier, dehumidifiers, media\_player \\ 
Dining Room & light, fan air\_purifiers, humidifier, dehumidifiers, trash\\ 
Study Room & light, air\_conditioner, heating, fan, air\_purifiers, humidifier, dehumidifiers, trash, aromatherapy\\ 
Kitchen & light, fan, trash, water\_heater\\ 
Bathroom & light, heating, trash, water\_heater\\ 
Balcony & light, aromatherapy, trash, media\_player\\ 
Store Room &  light, air\_purifiers, humidifier, dehumidifiers\\ 
\hline
\end{tabular}%
}
\caption{Distribution of devices across all rooms in the virtual environment.}
\label{table:room and devices}
\end{table*}

\begin{table*}[]
\centering
\resizebox{\textwidth}{!}{%
\begin{tabular}{lc}
\hline
\textbf{Device Name} & \textbf{Methods} \\ 
\hline
light & turn\_on; turn\_off; set\_brightness; set\_mode \\ 
air\_conditioner& turn\_on; turn\_off; set\_temperature; set\_mode; set\_fan\_speed; set\_swing; \\
heating        & turn\_on; turn\_off; set\_temperature; set\_mode; set\_fan\_speed            \\
fan & turn\_on; turn\_off; set\_speed; set\_swing \\
air\_purifiers   & turn\_on; turn\_off; set\_mode; set\_fan\_speed                            \\
water\_heater    & turn\_on; turn\_off; set\_temperature; set\_mode                            \\
media\_player    & turn\_on; turn\_off; play; pause; stop; set\_volume; set\_song; set\_artist; set\_style            \\
trash & pack\\
aromatherapy   & turn\_on; turn\_off; set\_intensity; set\_interval \\
humidifier     & turn\_on; turn\_off; set\_intensity; set\_mode                              \\
dehumidifiers  & turn\_on; turn\_off; set\_intensity; set\_mode                              \\ 
tv & turn\_on; turn\_off; set\_volume; change\_channel; \\
\hline
\end{tabular}%
}
\caption{List of devices and their associated executable methods in the virtual environment.}
\label{table:device and methods}
\end{table*}

\section{Dataset Details}
\subsection{Virtual Environment Details}
\label{sec:appendix_env}
This section provides comprehensive statistics on our virtual environment. Table \ref{table:room and devices} provides a detailed overview of all rooms in the virtual environment, along with the corresponding devices located within each room. Table \ref{table:device and methods} presents a complete list of devices and their associated executable methods, detailing the executable methods available for each device. These tables provide a comprehensive understanding of the virtual environment's layout and the operational interfaces of devices.

\subsection{Commands Generation Details}
\label{sec:appendix_cmd}
In this section, we detail the specific prompts used for guiding the LLMs to generate user commands. Table \ref{table:prompt_generate} presents the prompt designed for generating the complete and most clear (Le 1) user commands. Additionally, the system prompts used for generating higher-elliptical commands are provided in Table \ref{table:prompt_otherlevels}. We also provide an example of device methods description in the prompt in Table \ref{example of methods}.

\section{Experiments}
\label{appen_exp}
\subsection{Implementation Details of Baselines}
\label{baselines details}
For SAGE and Sasha, we use the framework provided by \citet{rivkin2024aiot}. Since SAGE and Sasha require the LLM to possess the capability of tool calling, and only Qwen2.5 supports native function-call ability among open-source LLMs, we conducted evaluations on the Qwen2.5 series.

\textbf{SAGE} As a central hub, SAGE interfaces with multiple specialized tools: smart\_device\_tool for device control, memory\_tool for command history, and weather\_tool for environmental data collection. The smart\_device\_tool, functioning as an independent intelligent agent, provides essential device management functions including get\_methods(), list\_devices(), list\_rooms(), and execute\_command(). Based on these multiple tools, SAGE dynamically generates prompt trees and toolchains to handle user commands. To optimize system performance, we have streamlined the toolset by removing non-essential components such as code tools and external interaction capabilities, thereby reducing potential interference with the model's decision-making process.   

\textbf{Sasha} An intelligent agent designed for smart home control. It communicates with devices using JSON format and integrates memory and environmental data to enhance device control capabilities. Sasha uses its plan generation tool to generate high-quality action plans for under-specified user commands. Once the agents complete their operations, we verify the success of the execution by checking the final state of the devices. In contrast with Sage, Sasha only has ReAct-style \cite{yao2023react} planner tool while Sage has multiple tools such as smart\_device\_tool, memory\_tool, and so on.

\begin{table*}[!ht]
\centering
\setlength{\tabcolsep}{5pt}
\small
\begin{tabular}{p{2.2cm}p{6cm}p{6cm}}
\toprule
\textbf{Error Type} & \textbf{User Command} & \textbf{Generated (red) / Golden (green)} \\
\midrule

Room Missing
& ``\textcolor{deeporange}{Lights on, cool white}, please! \dots''
& \textcolor{deepred}{light.turn\_on(),
  light.set\_mode(``cool white'')}

  \textcolor{deepgreen}{bathroom.light.turn\_on(), \newline bathroom.light.set\_mode(``cool white'')} \\

\midrule

Room Error
& ``Hey, can you handle the \textcolor{deeporange}{aromatherapy} setup? \dots''
& \textcolor{deepred}{living\_room.aromatherapy.turn\_on(),\newline
  living\_room.aromatherapy.set\_intensity(80)}

  \textcolor{deepgreen}{balcony.aromatherapy.turn\_on(),\newline
  balcony.aromatherapy.set\_intensity(0)} \\

\midrule

Parameter Missing
& ``Hey, could you manage that \textcolor{deeporange}{trash} thing we discussed?''
& \textcolor{deepred}{dining\_room.trash.pack()}

  \textcolor{deepgreen}{dining\_room.trash.pack(18:46)} \\

\midrule

Parameter Error
& ``Could you make sure the humidifier is off by \textcolor{deeporange}{11:30 PM}? \dots''
& \textcolor{deepred}{guest\_bedroom.humidifier\newline.turn\_off(``23:30'')}

  \textcolor{deepgreen}{guest\_bedroom.humidifier\newline.turn\_off(23:30)} \\

\midrule

Ignoring History
& ``Hey, can you handle that thing with the \textcolor{deeporange}{air conditioner}? \dots''

  \textit{History:} ``Could you have the air conditioning kick \dots''

  \textit{Hist.\ resp.:} \textcolor{deeporange}{guest\_bedroom\newline.air\_conditioner.turn\_on(07:13)}
& \textcolor{deepred}{guest\_bedroom.air\_conditioner\newline.turn\_on(07:00)}

  \textcolor{deepgreen}{guest\_bedroom.air\_conditioner\newline.turn\_on(07:13)} \\

\bottomrule
\end{tabular}
\caption{Five representative error types in \datasetname: Room Missing, Room Error,
Parameter Missing, Parameter Error, and Ignoring History. The core component mentioned
in user commands is shown in \textcolor{deeporange}{orange}, the error response generated
by LLMs is shown in \textcolor{deepred}{red}, and the golden answer is shown in
\textcolor{deepgreen}{green}.}
\label{table:error_analysis}
\end{table*}

\textbf{Zero-Shot Prompting} We utilize the same prompt as the In-Context Learning prompt but remove the examples and incorporate specific format requirements. In this approach, the model does not have access to the user's dialogue history.

\textbf{In-Context Learning} We provide the prompt used in In-Context Learning in Table \ref{table:prompt_ICL}. For each level of elliptical commands, we include one illustrative example. In this approach, the model does not have access to the user's dialogue history neither.

\textbf{RAG} We implement retrieval-augmented generation using vector database. All user commands and the corresponding generated device operations are stored in a database. The user's current instruction is used as a query to search through the dialogue history, with a fixed setting of top$k$ = 3 for all experiments. To evaluate the impact of preloaded irrelevant memory, we additionally collected fifty natural language instructions from users and randomly introduced 5, 10, and 20 of these as irrelevant memory.

\subsection{Metrics}
\label{app_metrics}
\textbf{F1} is calculated as follows:
\begin{equation}
    Precision=\frac{\text { operation\_correct\_num }}{\text {operation\_pred\_num }}
\end{equation}
\begin{equation}
    Recall=\frac{\text { operation\_correct\_num }}{\text {operation\_gold\_num }}
\end{equation}
\begin{equation} 
F1 = 2 \cdot \frac{Precision \cdot Recall}{Precision + Recall} \end{equation}

Here, Precision measures the proportion of correctly predicted operations out of all predicted operations, while Recall measures the proportion of correctly predicted operations out of all ground truth operations.

In contrast to F1, \textbf{Execution Accuracy (EA)} is measured by extracting the operations generated by the model without requiring the model's output format to be strictly correct. (In this paper, the model is required to enclose its generated instructions within \{\}; however, for EA calculation, we consider whether the generated operations are functionally correct, assuming format issues can be addressed as an engineering problem later.) For F1, we apply a stricter requirement that the format of the generated instructions must also be correct.

Execution Accuracy (EA) assesses the functional correctness of model-generated operations without enforcing strict output formatting. In this work, although models are instructed to enclose generated instructions within curly braces (\{\}), EA considers only whether the predicted operations would achieve the user intended operation execution in the environment, treating syntactic inconsistencies as engineering issues that can be addressed post hoc. In contrast, F1 applies a more strict criterion where outputs must not only be functional correct but also strictly adhere to the prescribed format. 

This two-tiered evaluation strategy enables a nuanced analysis by decoupling semantic understanding from surface-level formatting compliance. EA reflects the model’s core capacity to interpret natural language instructions and identify appropriate device operations, independent of format errors. Meanwhile, F1 captures practical deployability by measuring conformity to interface standards required for downstream execution. The combination of these metrics provides a more comprehensive assessment of the model’s core interpretation capabilities and practical utility, enabling evaluation of both its semantic understanding and its ability to generate outputs suitable for real-world deployment.

\subsection{Implementation Details}
The experiments with open-source models were conducted on NVIDIA A100 GPUs and NVIDIA A800 GPUs. For GPT-4o, we utilized the APIs provided by OpenAI. When fine-tuning Qwen2.5-7B-Instruct, we performed the fine-tuning experiments on NVIDIA A800 GPUs using the LoRA (Low-Rank Adaptation) \cite{hu2022lowrank} technique. The hyperparameters for LoRA are configured as follows: $r$=16, $lora\_alpha$=32, $learning\_rate$=1e-5.

\section{Error Analysis}
\label{Error Analysis}

We conduct a comprehensive error analysis to identify the challenges in interpreting progressively elliptical commands. Inspired by \citet{yin2024harmonyhomeagentresponsive, yu2018spider}, we categorize five representative error types, as shown in Table~\ref{table:error_analysis}. To ensure accuracy beyond automated metrics and provide a quantitative understanding, we conducted a manual root-cause analysis on 100 randomly sampled error cases from the Qwen2.5-7B (RAG) results. The distribution of the primary failure modes is detailed below:

\textbf{Ignoring History (38\%)}: 
The most frequent error. It occurs when the model fails to leverage dialogue history retrieved from the database. The model treats the elliptical command in isolation, failing to retrieve entities (e.g., air conditioner) or states established in previous turns.

\textbf{Room Missing (22\%)}: Occurs when user instructions specify a device’s operation but omit its location. The operation is correct, but the model fails to infer the spatial context (e.g., outputting \texttt{turn\_on} without specifying \texttt{bedroom}).
    
\textbf{Room Error (17\%)}: Occurs when the model infers the wrong room due to ambiguous context switching.
    
\textbf{Parameter Error (13\%)}: 
Involves incorrect parameter formatting or value assignment. While some type conversions (e.g., string to integer) might seem trivial, models often hallucinate formats (e.g., generating ``11:30 PM'' instead of the required ``23:30'') or misinterpret vague values (e.g., ``a bit warmer''). In strict API-based home automation, adhering to the exact schema is critical; a ``close enough'' parameter usually results in execution failure (API Error). Thus, strictly capturing these as errors is vital for evaluating an agent's robustness.
    
\textbf{Parameter Missing (10\%)}: Results from incomplete parameter extraction for device operations.

It is important to note that structural formatting errors are not separately categorized in this analysis. This is because, during the Execution Accuracy (EA) calculation, models' responses that are operationally correct but have syntax-level formatting errors are still considered correct. Such common structural errors include missing the required curly braces \{\} for function-call style operations, or the erroneous use of other delimiters such as single quotes \texttt{` '} or angle brackets \textless\textgreater. This leniency strictly applies to the outer syntax, whereas API parameter values must remain precise as discussed above.

\input{latex/tables/prompt_generate}

\input{latex/tables/prompt_otherlevels}

\input{latex/tables/prompt_ICL}

\begin{table*}[!ht]
\centering
\begin{tcolorbox}[
    colback=gray!10,
    colframe=black,
    width=14cm,
    arc=2mm,
    auto outer arc,
    title={Example}
]
light

light.turn\_on(time: Optional[str] = None, format: `\%H:\%M');

light.turn\_off(time: Optional[str] = None, format: `\%H:\%M');

light.set\_brightness(brightness: int) (range: 0 -- 4);

light.set\_mode(mode: str) (options: [``soft warm'', ``neutral'', ``cool white'', ``daylight'', ``cool'']);

\end{tcolorbox}
\caption{Examples of home device methods.}
\label{example of methods}
\end{table*}

\begin{figure}[!ht]
    \centering
    \includegraphics[width=\linewidth]{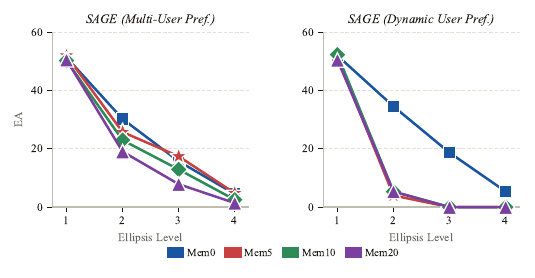}
    \caption{Execution Accuracy of SAGE on Qwen2.5-7B-Instruct across varying amounts of preloaded memory in multi-user preferences and dynamic user preferences scenarios. `Mem number' indicates the amount of preloaded irrelevant memory.}
\label{SAGE_mem}
\end{figure}

\begin{figure}[!ht]
    \centering
    \includegraphics[width=\linewidth]{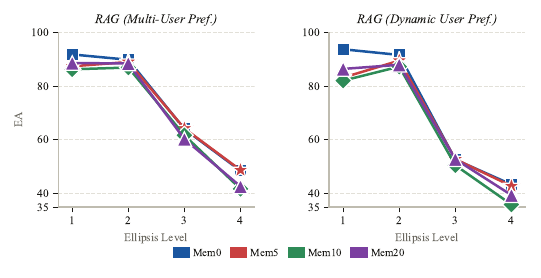}
    \caption{Execution Accuracy of RAG on Qwen2.5-7B-Instruct across varying amounts of preloaded memory in multi-user preferences and dynamic user preferences scenarios. `Mem number' indicates the amount of preloaded irrelevant memory.} 
\label{RAG_mem}
\end{figure}

\begin{figure}[!ht]
    \centering
    \includegraphics[width=\linewidth]{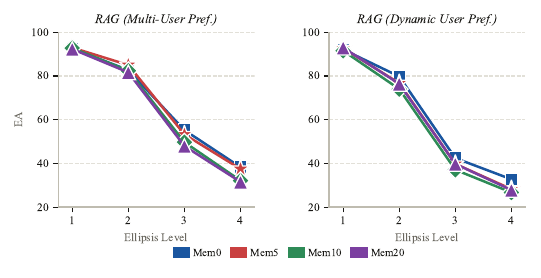}
    \caption{Execution Accuracy of RAG on Gemma2-9b-it across varying amounts of preloaded memory in multi-user preferences and dynamic user preferences scenarios. `Mem number' indicates the amount of preloaded irrelevant memory.}
\label{RAG_mem_gemma}
\end{figure}

\begin{figure}[!ht]
    \centering
    \includegraphics[width=\linewidth]{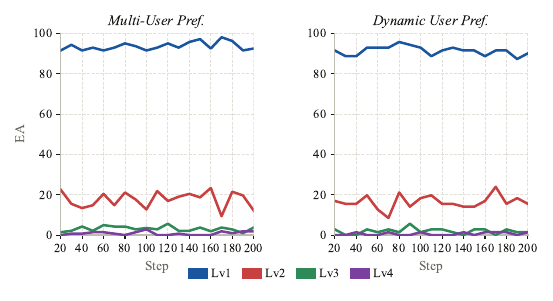}
    \caption{Execution Accuracy of Gemma2-9b-it model in test dataset under different training steps.}
    \label{ft_gemma}
\end{figure}

\end{document}

%% file: latex/tables/comparsion.tex
\begin{table*}[!ht]
\centering
    \tabcolsep=0.13cm
    \small
    \begin{tabular*}{\hsize}{@{}@{\extracolsep{\fill}}cccccccc@{}}
\toprule
    \textbf{Dataset} & \textbf{Prog. Ellipsis} & \textbf{Multi Pref.} & \textbf{Dynamic Pref.} & \textbf{Long-Inter.} & \textbf{Persona} & \textbf{Pers. Size} & \textbf{Total}
    \\
\midrule
    \textbf{IFTTT \cite{Yu_2021}} 
    & \color{deepred}\XSolidBrush 
    & \color{deepred}\XSolidBrush 
    & \color{deepred}\XSolidBrush 
    & \color{deepred}\XSolidBrush 
    & \color{deepred}\XSolidBrush
    & 0 & 50,000+$^*$
    \\
    
    \textbf{Sasha \cite{10.1145/3643505}} 
    & \color{deepred}\XSolidBrush 
    & \color{deepred}\XSolidBrush 
    & \color{deepred}\XSolidBrush 
    & \color{deepred}\XSolidBrush
    & \color{deepred}\XSolidBrush
    & 0 & 60
    \\

    \textbf{SAGE \cite{rivkin2024aiot}} 
    & \color{deepred}\XSolidBrush 
    & \color{deepred}\XSolidBrush 
    & \color{deepred}\XSolidBrush 
    & \color{deepred}\XSolidBrush
    & \color{deepgreen}\CheckmarkBold
    & 3 & 50
    \\

\midrule
    \textbf{\datasetname(Ours)} 
    & \color{deepgreen}\CheckmarkBold  
    & \color{deepgreen}\CheckmarkBold 
    & \color{deepgreen}\CheckmarkBold 
    & \color{deepgreen}\CheckmarkBold 
    & \color{deepgreen}\CheckmarkBold 
    & 1,424 & 7,120
    \\ 
\bottomrule
    \end{tabular*}
    \caption{Comparison of existing simulated home datasets. Key features include Progressively Elliptical Commands (Prog. Ellipsis), Multi-User Preferences (Multi Pref.), Dynamic User Preferences (Dynamic Pref.), Long-Term Interaction Support (Long-Inter.), and Persona Size (Pers. Size). $^*$ The IFTTT dataset uses a hard-coded `If This Then That' format, which differs from the natural language instructions used in the other datasets (Sasha, SAGE, \datasetname).}
    \label{table:data_comparison}
\end{table*}

%% file: latex/tables/statistics.tex
\begin{table*}[t]
    \centering
    \renewcommand{\arraystretch}{0.8}
    \begin{tabular*}{\hsize}{@{}@{\extracolsep{\fill}}lcccccccc@{}}
    \toprule
    & \multicolumn{4}{c}{\textbf{Multi-User Preferences}} & \multicolumn{4}{c}{\textbf{Dynamic User Preferences}} \\
    \cmidrule(lr){2-5} \cmidrule(lr){6-9}
    \textbf{Statistics} & Lv1 & Lv2 & Lv3 & Lv4 & Lv1 & Lv2 & Lv3 & Lv4 \\
    \midrule
    Avg. Tokens          & 42.08 & 34.54 & 33.88 & 28.43 & 42.10 & 34.70 & 33.87 & 28.11 \\
    Num. Commands        & 1,068 & 1,068 & 1,068 & 1,068 & 712 & 712 & 712 & 712 \\
    Room (\%)            & 99.34 & 1.59  & 1.03  & 12.36 & 99.02 & 1.26  & 0.98  & 11.10 \\
    Device (\%)          & 98.97 & 95.69 & 82.30 & 91.29 & 98.60 & 96.63 & 79.92 & 91.57 \\
    Operation (\%)       & 91.48 & 79.49 & 26.40 & 4.49  & 90.73 & 77.25 & 26.54 & 6.04  \\
    \bottomrule
    \end{tabular*}
    \caption{Statistics of \datasetname~across different ellipsis levels (Lv1-Lv4) for two tasks: Multi-User Preferences and Dynamic User Preferences. Core component percentages reflect the presence of Room, Device, and Operation specifications in user commands.}
    \label{table:Statistics}
\end{table*}

%% file: latex/tables/main.tex
\begin{table*}[!t]
\centering
\small
\renewcommand{\arraystretch}{1.0}
\setlength{\tabcolsep}{1mm}
\definecolor{lightgray}{RGB}{240,240,240}
\begin{tabular}{lcccccccccccccccc}
\toprule
\multicolumn{1}{l}{} & \multicolumn{8}{c}{\textbf{Multi-User Preferences}} & \multicolumn{8}{c}{\textbf{Dynamic User Preferences}} \\ 
\cmidrule(lr){2-9} \cmidrule(lr){10-17}
\multicolumn{1}{l}{} & \multicolumn{2}{c}{\textbf{Lv1}} & \multicolumn{2}{c}{\textbf{Lv2}} & \multicolumn{2}{c}{\textbf{Lv3}} & \multicolumn{2}{c}{\textbf{Lv4}} & \multicolumn{2}{c}{\textbf{Lv1}} & \multicolumn{2}{c}{\textbf{Lv2}} & \multicolumn{2}{c}{\textbf{Lv3}} & \multicolumn{2}{c}{\textbf{Lv4}} \\ 
\cmidrule(lr){2-3} \cmidrule(lr){4-5} \cmidrule(lr){6-7} \cmidrule(lr){8-9}
\cmidrule(lr){10-11} \cmidrule(lr){12-13} \cmidrule(lr){14-15} \cmidrule(lr){16-17}
\multicolumn{1}{l}{\multirow{-3}{*}{\textbf{Method}}} & \textbf{EA} & \textbf{F1} & \textbf{EA} & \textbf{F1} & \textbf{EA} & \textbf{F1} & \textbf{EA} & \textbf{F1} & \textbf{EA} & \textbf{F1} & \textbf{EA} & \textbf{F1} & \textbf{EA} & \textbf{F1} & \textbf{EA} & \textbf{F1} \\ 
\midrule
\multicolumn{17}{c}{\textbf{LLaMA3-8B}} \\ \midrule
\rowcolor{lightgray}
\multicolumn{1}{l}{0-Shot} & 45.04 & 38.94 & 7.87 & 6.80 & 1.12 & 1.94 & 0.56 & 2.21 & 36.66 & 29.42 & 6.32 & 5.83 & 0.56 & 0.95 & 0.28 & 0.96 \\
\multicolumn{1}{l}{ICL} & 79.21 & 75.57 & 10.49 & 10.71 & 2.81 & 4.29 & 1.40 & 5.49 & 74.58 & 72.18 & 7.87 & 9.23 & 1.40 & 2.94 & 0.84 & 3.93 \\
\rowcolor{lightgray}
\multicolumn{1}{l}{RAG} & 73.97 & 72.04 & 71.82 & 69.97 & 58.71 & 58.41 & 42.88 & 43.19 & 74.30 & 66.73 & 58.29 & 58.54 & 35.67 & 38.46 & 24.02 & 28.51 \\ \midrule
\multicolumn{17}{c}{\textbf{Mistral-7B}} \\ \midrule
\rowcolor{lightgray}
\multicolumn{1}{l}{0-Shot} & 58.43 & 47.68 & 6.84 & 5.98 & 1.22 & 1.06 & 1.31 & 1.56 & 57.16 & 40.24 & 6.60 & 5.03 & 1.26 & 0.49 & 0.56 & 0.62 \\
\multicolumn{1}{l}{ICL} & 90.36 & 77.77 & 16.11 & 15.63 & 2.81 & 3.79 & 0.66 & 4.57 & 88.76 & 79.32 & 19.52 & 17.83 & 2.25 & 4.59 & 1.12 & 4.89 \\
\rowcolor{lightgray}
\multicolumn{1}{l}{RAG} & 88.76 & 83.46 & 83.15 & 79.92 & 63.76 & 62.20 & 47.47 & 49.75 & 79.92 & 73.05 & 70.93 & 67.44 & 44.24 & 44.94 & 33.99 & 37.51 \\ \midrule
\multicolumn{17}{c}{\textbf{Gemma2-9B}} \\ \midrule
\rowcolor{lightgray}
\multicolumn{1}{l}{0-Shot} & 48.41 & 51.76 & 8.33 & 9.30 & 0.84 & 3.36 & 0.28 & 3.43 & 44.94 & 50.62 & 8.15 & 9.02 & 0.56 & 1.82 & 0.28 & 3.26 \\
\multicolumn{1}{l}{ICL} & 93.26 & 75.33 & 16.20 & 13.18 & 2.53 & 3.63 & 0.66 & 4.25 & 91.85 & 77.55 & 15.87 & 12.57 & 1.97 & 3.73 & 0.56 & 4.57 \\
\rowcolor{lightgray}
\multicolumn{1}{l}{RAG} & 92.60 & 90.76 & 82.68 & 82.24 & 55.24 & 59.75 & 38.30 & 47.73 & 92.13 & \cellcolor[HTML]{BEBEF4}89.89 & 79.78 & 77.19 & 42.42 & 47.62 & 32.72 & 42.69 \\ \midrule
\multicolumn{17}{c}{\textbf{Gemma2-27B}} \\ \midrule
\rowcolor{lightgray}
\multicolumn{1}{l}{0-Shot} & 48.60 & 57.06 & 6.93 & 7.93 & 1.03 & 2.77 & 0.94 & 2.70 & 45.51 & 47.27 & 7.44 & 8.22 & 0.56 & 1.47 & 0.28 & 2.40 \\
\multicolumn{1}{l}{ICL} & 93.63 & 77.80 & 15.36 & 14.02 & 2.81 & 4.82 & 0.84 & 5.25 & 91.01 & 78.25 & 13.90 & 11.01 & 1.97 & 3.21 & 0.84 & 4.32 \\
\rowcolor{lightgray}
\multicolumn{1}{l}{RAG} & 91.95 & 78.46 & 87.36 & 82.58 & 62.27 & 65.60 & 45.04 & 53.14 & 92.98 & 76.98 & \cellcolor[HTML]{BEBEF4}88.37 & 78.59 & 57.87 & 59.55 & 43.12 & 51.06 \\ \midrule
\multicolumn{17}{c}{\textbf{Qwen2.5-7B}} \\ \midrule
\rowcolor{lightgray}
\multicolumn{1}{l}{0-Shot} & 54.59 & 55.00 & 9.46 & 9.65 & 0.94 & 2.43 & 0.66 & 3.16 & 54.21 & 45.44 & 11.38 & 9.76 & 0.70 & 1.03 & 0.28 & 1.51 \\
\multicolumn{1}{l}{ICL} & 88.58 & 83.95 & 15.45 & 12.42 & 2.62 & 3.61 & 1.22 & 4.85 & 86.94 & 80.10 & 17.42 & 13.69 & 2.25 & 3.85 & 1.40 & 4.89 \\
\rowcolor{lightgray}
\multicolumn{1}{l}{RAG} & 91.76 & 87.25 & 89.79 & 85.18 & 64.04 & 65.05 & 48.50 & 53.94 & \cellcolor[HTML]{F4BEBE}93.68 & 86.88 & \cellcolor[HTML]{F4BEBE}91.57 & \cellcolor[HTML]{F4BEBE}{\color[HTML]{333333} 85.97} & 52.67 & 57.17 & 43.40 & 50.56 \\ \midrule
\multicolumn{17}{c}{\textbf{Qwen2.5-14B}} \\ \midrule
\rowcolor{lightgray}
\multicolumn{1}{l}{0-Shot} & 50.94 & 58.26 & 8.71 & 9.50 & 2.06 & 3.38 & 1.31 & 3.60 & 38.48 & 48.35 & 8.00 & 9.71 & 0.98 & 2.97 & 0.84 & 3.36 \\
\multicolumn{1}{l}{ICL} & 85.86 & 74.83 & 10.67 & 10.87 & 2.62 & 3.98 & 1.69 & 4.79 & 81.88 & 71.18 & 12.08 & 12.68 & 2.67 & 4.14 & 1.97 & 3.98 \\
\rowcolor{lightgray}
\multicolumn{1}{l}{RAG} & 88.11 & 88.79 & 86.61 & 87.13 & 69.19 & \cellcolor[HTML]{BEBEF4}73.50 & 48.41 & \cellcolor[HTML]{BEBEF4}57.86 & 79.21 & 81.65 & 66.57 & 68.68 & 47.33 & 53.79 & 40.73 & 49.33 \\ \midrule
\multicolumn{17}{c}{\textbf{Qwen2.5-32B}} \\ \midrule
\rowcolor{lightgray}
\multicolumn{1}{l}{0-Shot} & 55.06 & 61.27 & 10.21 & 11.98 & 2.15 & 5.24 & 1.12 & 4.75 & 50.14 & 55.97 & 9.27 & 11.19 & 1.40 & 3.73 & 1.12 & 4.33 \\
\multicolumn{1}{l}{ICL} & 92.13 & 89.89 & 14.23 & 14.14 & 2.15 & 4.52 & 1.69 & 4.22 & 88.20 & 85.35 & 6.46 & 5.93 & 1.55 & 2.20 & 1.26 & 2.91 \\
\rowcolor{lightgray}
\multicolumn{1}{l}{RAG} & 89.14 & 84.01 & 87.92 & 75.08 & 68.07 & 58.54 & 48.97 & 47.68 & 90.31 & 84.01 & 79.92 & 75.08 & 58.43 & 58.54 & 43.68 & 47.68 \\ \midrule
\multicolumn{17}{c}{\textbf{Qwen2.5-72B}} \\ \midrule
\rowcolor{lightgray}
\multicolumn{1}{l}{0-Shot} & 39.32 & 51.76 & 7.68 & 9.87 & 2.62 & 4.67 & 1.97 & 3.83 & 34.27 & 47.82 & 7.16 & 9.61 & 2.39 & 4.44 & 1.69 & 3.74 \\
\multicolumn{1}{l}{ICL} & 89.42 & 88.11 & 15.72 & 13.17 & 4.03 & 4.41 & 2.25 & 4.46 & 88.90 & 75.76 & 16.29 & 15.90 & 2.81 & 4.58 & 2.11 & 3.91 \\
\rowcolor{lightgray}
\multicolumn{1}{l}{RAG} & \cellcolor[HTML]{F4BEBE}95.51 & \cellcolor[HTML]{F4BEBE}92.47 & 92.32 & \cellcolor[HTML]{BEBEF4}89.48 & 69.10 & 70.27 & 52.25 & 57.73 & 84.55 & 85.43 & 79.21 & \cellcolor[HTML]{BEBEF4}80.55 & 54.49 & \cellcolor[HTML]{BEBEF4}61.41 & 45.65 & \cellcolor[HTML]{BEBEF4}55.07 \\ \midrule
\multicolumn{17}{c}{\textbf{GPT-4o}} \\ \midrule
\rowcolor{lightgray}
\multicolumn{1}{l}{0-Shot} & 55.99 & 61.57 & 5.06 & 6.20 & 1.03 & 3.07 & 1.31 & 2.68 & 52.81 & 65.14 & 5.62 & 7.21 & 1.12 & 4.03 & 0.70 & 4.63 \\
\multicolumn{1}{l}{ICL} & 92.98 & 90.25 & 14.75 & 14.94 & 3.37 & 5.74 & 1.40 & 5.25 & \cellcolor[HTML]{BEBEF4}93.26 & \cellcolor[HTML]{F4BEBE}92.02 & 17.98 & 17.86 & 3.46 & 5.85 & 1.69 & 6.18 \\
\rowcolor{lightgray}
\multicolumn{1}{l}{RAG} & 93.07 & \cellcolor[HTML]{BEBEF4}90.98 & \cellcolor[HTML]{BEBEF4}92.70 & \cellcolor[HTML]{F4BEBE}90.72 & \cellcolor[HTML]{BEBEF4}76.05 & \cellcolor[HTML]{F4BEBE}77.69 & \cellcolor[HTML]{F4BEBE}55.71 & \cellcolor[HTML]{F4BEBE}64.21 & 91.57 & 82.23 & 88.34 & 78.97 & \cellcolor[HTML]{BEBEF4}61.24 & \cellcolor[HTML]{F4BEBE}{\color[HTML]{333333} 65.53} & \cellcolor[HTML]{F4BEBE}50.70 & \cellcolor[HTML]{F4BEBE}58.47 \\ \midrule
\multicolumn{17}{c}{\textbf{DeepSeek-V3}} \\ \midrule
\rowcolor{lightgray}
\multicolumn{1}{l}{0-Shot} & 55.15 & 62.58 & 11.24 & 13.00 & 2.81 & 4.72 & 2.06 & 5.18 & 52.81 & 59.97 & 6.74 & 8.69 & 1.40 & 2.59 & 1.69 & 3.37 \\
\multicolumn{1}{l}{ICL} & 92.79 & 89.42 & 13.02 & 11.84 & 3.18 & 4.41 & 1.78 & 5.13 & 92.84 & 88.43 & 4.78 & 4.85 & 1.69 & 2.01 & 0.98 & 4.55 \\
\rowcolor{lightgray}
\multicolumn{1}{l}{RAG} & \cellcolor[HTML]{BEBEF4}94.10 & 81.72 & \cellcolor[HTML]{F4BEBE}92.88 & 80.46 & \cellcolor[HTML]{F4BEBE}76.78 & 70.57 & \cellcolor[HTML]{BEBEF4}53.37 & 54.68 & 80.62 & 62.91 & 78.65 & 61.71 & \cellcolor[HTML]{F4BEBE}62.08 & 52.52 & \cellcolor[HTML]{BEBEF4}49.58 & 46.12 \\ \bottomrule
\end{tabular}
\caption{Performance comparison across ellipsis levels for Multi-User Preferences and Dynamic User Preferences. Results are evaluated using Execution Accuracy (EA) and F1 Score (F1), where \colorbox{myorange}{orange} indicates the best performance and \colorbox{mypurple}{purple} denotes the second-best performance. 0-Shot, ICL, and RAG refer to Zero-Shot Prompting, In-Context Learning, and Retrieval-Augmented Generation, respectively.}
\label{table:main_result}
\end{table*}

%% file: latex/tables/prompt_generate.tex
\definecolor{deeporange}{RGB}{230, 81, 0}
\definecolor{darkblue}{RGB}{13, 71, 161}

\begin{table*}[!ht]
\centering
\begin{tcolorbox}[
    colback=gray!10,
    colframe=black,
    width=14cm,
    arc=2mm,
    auto outer arc,
    title={Prompt},
    breakable,
    enhanced jigsaw,
    before upper={\parindent15pt\noindent}
]
Act as a homeowner interacting casually with a smart home assistant.
Your task is to generate natural language commands that you would use to instruct your virtual assistant.
\textcolor{deeporange}{A clear command should include the device, operation, room, and trigger time or setting
and you should ensure that the command you generated should have those four parts.
You should incorporate the environment parameters, and the generated command should
correlate with the environment.}
Remember, you are the user issuing commands, not the assistant responding.
Ensure your commands are varied and reflect how a real person would naturally communicate
with their virtual assistant. You just need to generate one command.

\textcolor{darkblue}{\textless examples\textgreater}

\textcolor{darkblue}{Here are some examples:}

\textcolor{darkblue}{device : dehumidifier}

\textcolor{darkblue}{required instruction : \textless turn\_on(06:53)\textgreater}

\textcolor{darkblue}{given persona : \textless my mom is my best friend. I have four sisters.
I believe that mermaids are real. i love iced tea.\textgreater}

\textcolor{darkblue}{environment parameters : \textless ``humidity'': 70\textgreater}

\textcolor{darkblue}{respond content: \textless ``Hey, can you turn on the dehumidifier in the living room
at 6:53, please? It's getting a bit too humid in here. Thanks!''\textgreater}

\dots

\textcolor{darkblue}{\textless /examples\textgreater}

required instruction :

given persona :

environment parameters :

respond content :

\end{tcolorbox}
\caption{The prompt used to guide LLM to generate the most clear (Lv1) user commands.
The illustration of the manually defined ellipsis level is shown in \textcolor{deeporange}{orange} and a few shots are shown in \textcolor{darkblue}{blue}.}
\label{table:prompt_generate}
\end{table*}

%% file: latex/tables/prompt_otherlevels.tex
\begin{table*}[!ht]
\centering
\begin{tcolorbox}[
    colback=gray!10,
    colframe=black,
    width=14cm,
    arc=2mm,
    auto outer arc,
    title={Prompt},
    breakable,
    enhanced jigsaw,
    before upper={\parindent15pt\noindent}
]
The system prompt used for generating Lv2 user commands:

Act as a homeowner interacting casually with a smart home assistant.
You have given a full command, specifying the operation, device, and room.
\textcolor{deeporange}{Then, follow up with shorter commands that assume context, using natural
language and varying sentence structures. You should exclude the room in your commands,
while still specifying the operation, device, and trigger time/settings.}
The chat history is provided below. Remember, you are the user issuing commands, not the
assistant responding. Ensure your commands are varied and reflect how a real person would
naturally communicate with their virtual assistant. You just need to generate one command.

\pagebreak
\vspace{0.5cm}
The system prompt used for generating Lv3 user commands:

\dots\ \textcolor{deeporange}{The third command should be a more vague version compared to the
second command, using natural language and varied sentence structures. In that command, you
should not include the room, and ensure that the trigger time or setting remains ambiguous.
Additionally, make the operation itself somewhat ambiguous, based on the second command.}
Assume the assistant has interacted with the user multiple times and has learned their
preferences and habits. The chat history is provided below. \dots

\vspace{0.5cm}
The system prompt used for generating Lv4 user commands:

\dots\ \textcolor{deeporange}{The fourth command should be the most vague version compared to the others, using natural language and varied sentence structures. This time, you should avoid including the operation, room, or trigger time/setting in the command, and only reference the device.} Assume the assistant has interacted with the user multiple times and has learned their
preferences and habits. The chat history is provided below. \dots

\end{tcolorbox}
\caption{The prompts used to guide LLM to generate user commands of other ellipsis levels.
The illustration of manually defined ellipsis levels is shown in \textcolor{deeporange}{orange}.}
\label{table:prompt_otherlevels}
\end{table*}

%% file: latex/tables/prompt_ICL.tex
\begin{table*}[!ht]
\centering
\begin{tcolorbox}[
    colback=gray!10,
    colframe=black,
    width=14cm,
    arc=2mm,
    auto outer arc,
    title={Prompt},
    before upper={\parindent15pt\noindent}
]
You are a helpful AI Assistant that controls the devices in a house. Complete the following
task as instructed or answer the following question with the information provided only.
The devices and the methods devices possess are provided below, please only use the methods
provided. Only output assistant instructions and enclose them in \{\}. Please ensure that any
parameters involving time are expressed in a 24-hour format. For example: Use 14:00 to
represent 2:00 PM, not 2:00 PM. Use 08:30 to represent 8:30 AM, not 8:30 AM.

\textless device\_method\textgreater

The following provides the methods to control each device in the current house.
\dots\ (Device methods, Table~\ref{example of methods} shows an example.)

\textless /device\_method\textgreater

\textcolor{darkblue}{\textless example\textgreater}

\textcolor{darkblue}{Here are a few examples, your output format should be consistent with
the results provided in the example:}

\textcolor{darkblue}{user\_instruction: ``Hey, can you turn on the air conditioner in the
study room and set the temperature to 23 degrees? I'm getting ready to finalize some
paperwork from my recent fair, and I need a comfortable spot to focus.''}

\textcolor{darkblue}{assistant\_instruction: study\_room.air\_conditioner.turn\_on(),}

\textcolor{darkblue}{study\_room.air\_conditioner.set\_temperature(23)}

\dots

\textcolor{darkblue}{\textless /example\textgreater}

\textless environment\textgreater

The following provides the environment information of the current room.

\textless /environment\textgreater

\noindent\rule{\linewidth}{0.4pt}

Here are the user instructions you need to reply to.

user instructions :

assistant\_instruction :

\end{tcolorbox}
\caption{The prompt used to guide LLM to generate device operation using In-Context Learning.
Few shots are shown in \textcolor{darkblue}{blue}.}
\label{table:prompt_ICL}
\end{table*}